\documentclass[10pt,onecolumn,letterpaper]{article}

\usepackage{amsmath}
\usepackage{amssymb}
\usepackage{mathtools}
\usepackage{amsthm}
\usepackage{amsfonts,mathrsfs}
\usepackage{algorithm}
\usepackage{algpseudocode}
\theoremstyle{plain}
\newtheorem{theorem}{Theorem}[section]

\theoremstyle{definition}
\newtheorem{definition}[theorem]{Definition}

\theoremstyle{remark}

\usepackage{comment}

\newcommand\R{\mathbb{R}}

\usepackage[pagenumbers]{cvpr} 

\usepackage{ifthen}
\newboolean{show_main}
\newboolean{show_supp}

\definecolor{cvprblue}{rgb}{0.21,0.49,0.74}
\usepackage[pagebackref,breaklinks,colorlinks,allcolors=cvprblue]{hyperref}

\def\paperID{3450} 
\def\confName{CVPR}
\def\confYear{2025}

\title{Preconditioners for the Stochastic Training of Neural Fields}

\author{Shin-Fang Chng*
\hspace{1em}
Hemanth Saratchandran*
\hspace{1em}
Simon Lucey \\
Australian Institute for Machine Learning, University of Adelaide.  \\
}

\begin{document}

\twocolumn[{%
\renewcommand\twocolumn[1][]{#1}%
\maketitle

\vspace{-1.04cm}
\begin{center}
\centering
\captionsetup{type=figure}
\begin{subfigure}{0.32\linewidth} \includegraphics[width=\linewidth]{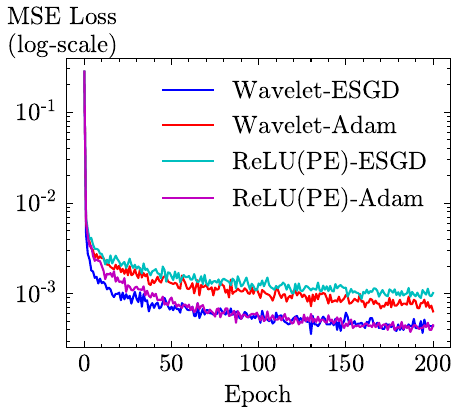}\caption{}\label{fig:1:wavelet-image}\end{subfigure}
\begin{subfigure}{0.32\linewidth} \includegraphics[width=\linewidth]{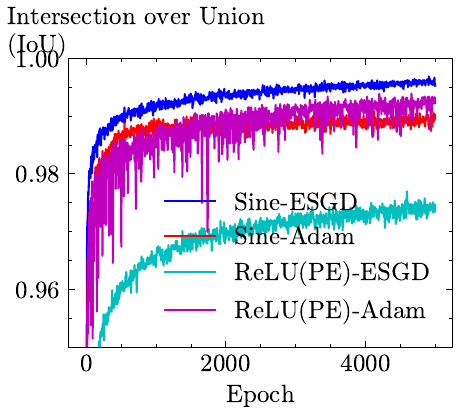}\caption{}\label{fig:1:sine-occupancy}\end{subfigure}
\begin{subfigure}{0.33\linewidth} \includegraphics[width=\linewidth]{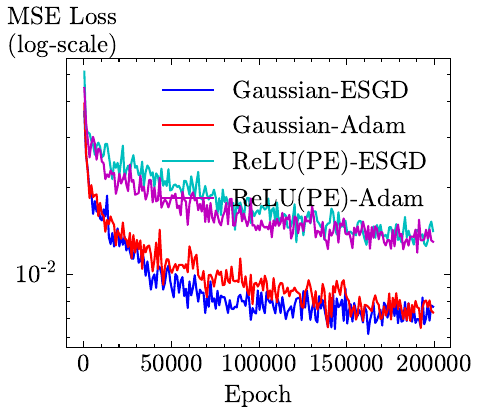}\caption{}\label{fig:1:gauss-nerf}\end{subfigure}
\begin{subfigure}{0.16\linewidth} \includegraphics[width=\linewidth]{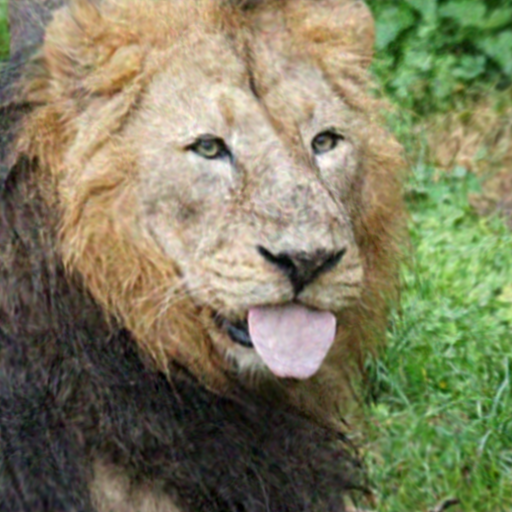}\caption{}\label{fig:1:wavelet-adam}\end{subfigure}
\begin{subfigure}{0.16\linewidth} \includegraphics[width=\linewidth]{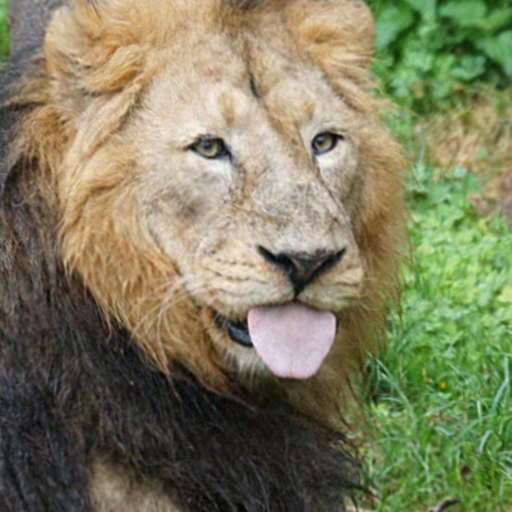}\caption{}\label{fig:1:wavelet-esgd}\end{subfigure}
\begin{subfigure}{0.16\linewidth} \includegraphics[width=\linewidth]{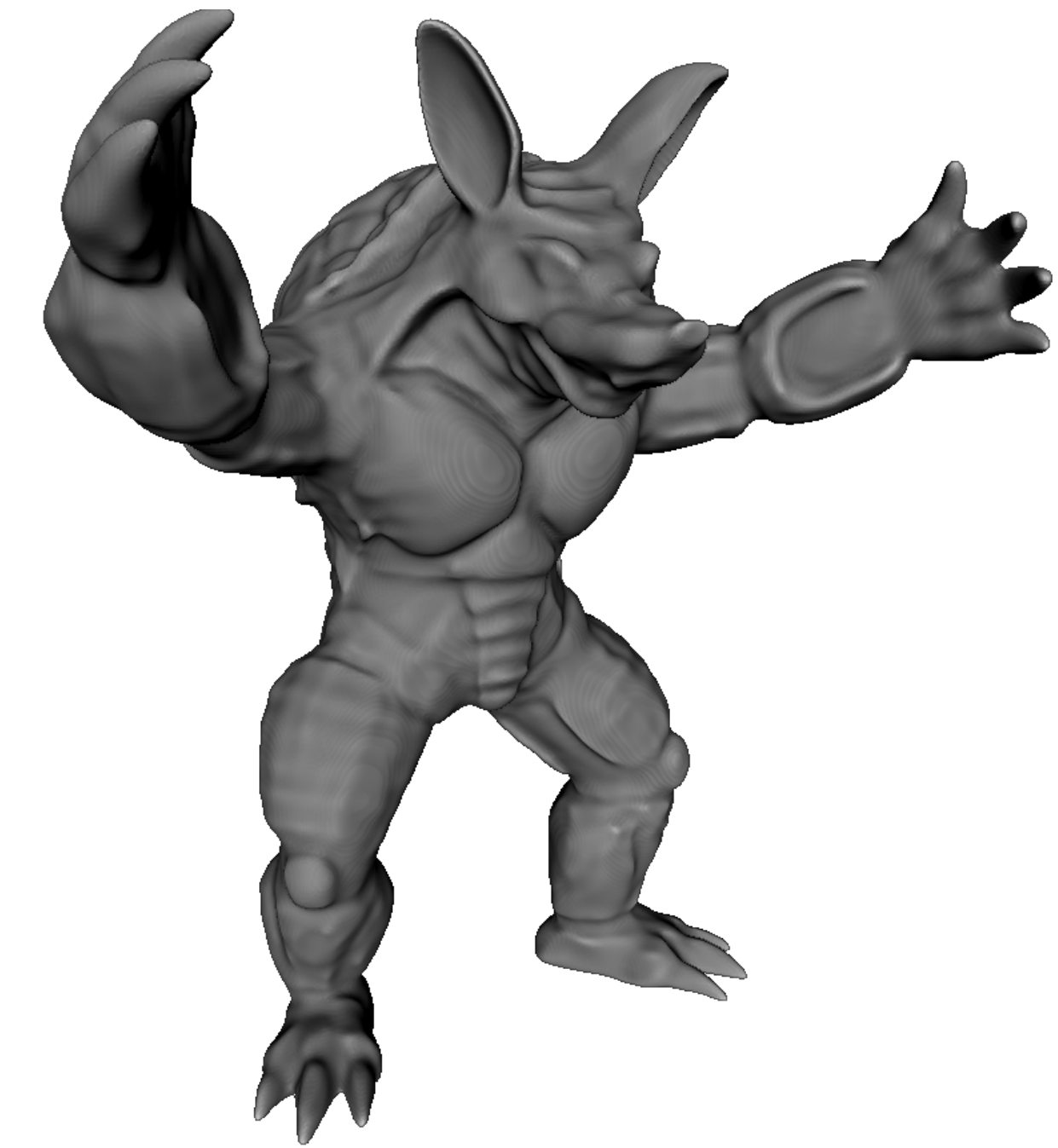}\caption{}\label{fig:1:shape-siren-adam}\end{subfigure}
\begin{subfigure}{0.16\linewidth} \includegraphics[width=\linewidth]{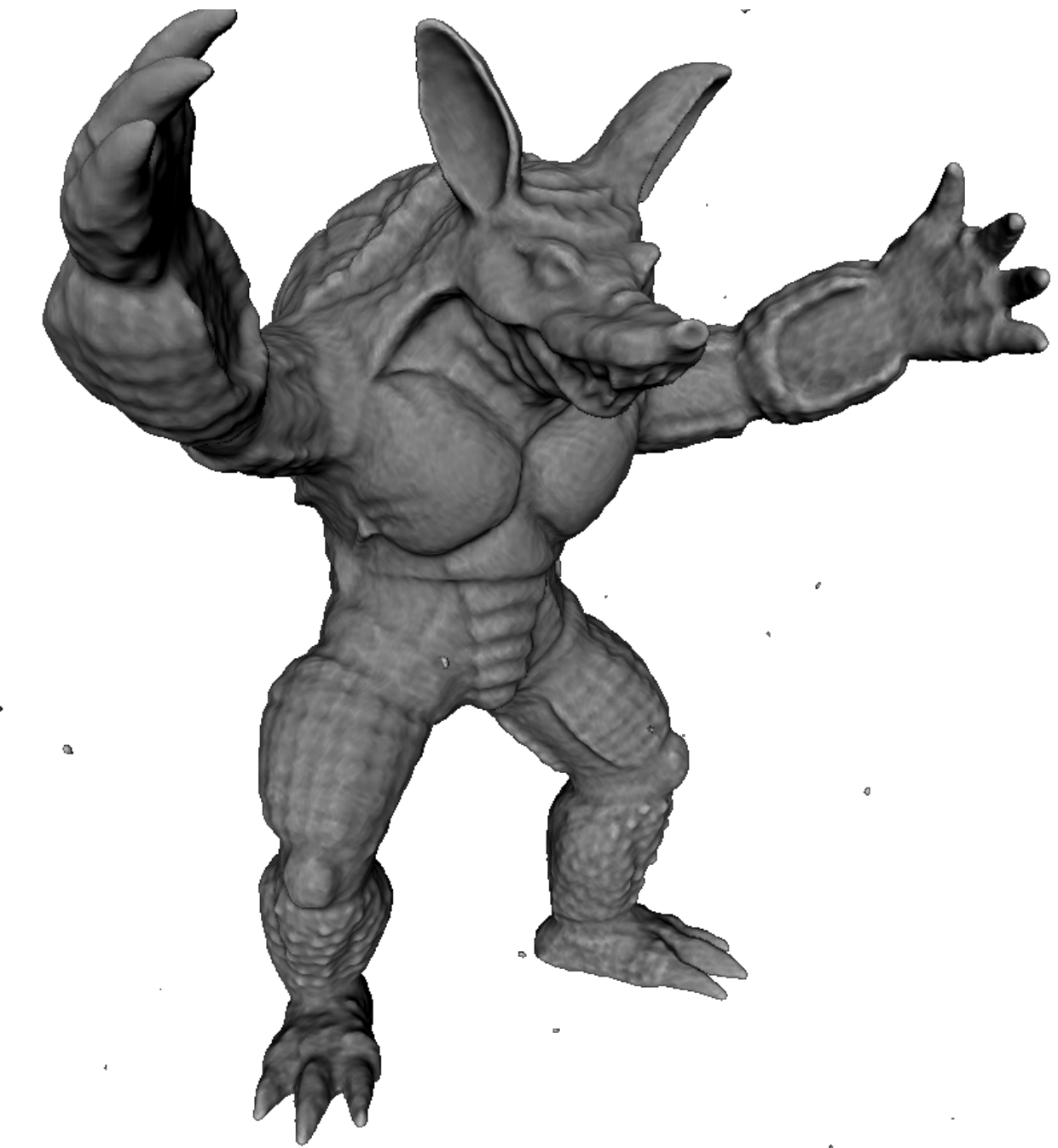}\caption{}\label{fig:1:shape-siren-esgd}\end{subfigure}
\begin{subfigure}{0.16\linewidth} \includegraphics[width=\linewidth]{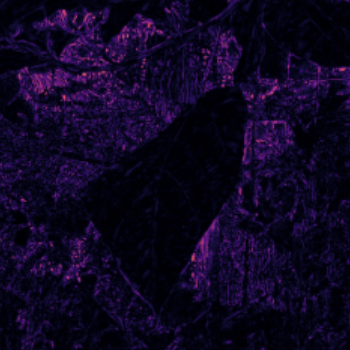}\caption{}\label{fig:1:nerf-gauss-adam}\end{subfigure}
\begin{subfigure}{0.16\linewidth} \includegraphics[width=\linewidth]{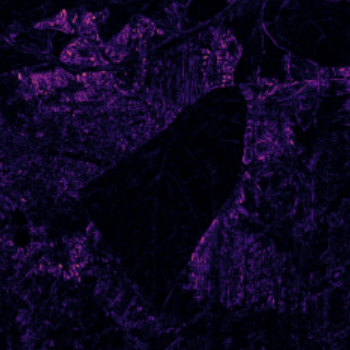}\caption{}\label{fig:1:nerf-gauss-esgd}\end{subfigure}
\vspace{-1em}
\caption{ \textbf{We demonstrate that curvature-aware preconditioners can significantly accelerate training convergence for neural fields with suitable activations}. Specifically, ESGD (a curvature-aware preconditioned gradient descent algorithm \cite{dauphin2015equilibrated}) improves convergence for wavelet, sine, and Gaussian activations, while Adam performs better for ReLU network with positional encoding (ReLU(PE)). In a \textit{2D image reconstruction} task (\cref{fig:1:wavelet-image}), a wavelet-based neural field trained with ESGD achieves \textcolor{blue}{29.36} PSNR (\cref{fig:1:wavelet-esgd}) at the $10^\text{th}$ epoch, outperforming Adam’s \textcolor{red}{26.61} PSNR (\cref{fig:1:wavelet-adam}). In a \textit{3D binary occupancy} task (\cref{fig:1:sine-occupancy}), ESGD enables faster convergence for sine activations, as shown by the improved armadillo mesh reconstruction at the $1000^\text{th}$ epoch (\cref{fig:1:shape-siren-esgd} vs. \cref{fig:1:shape-siren-adam}). In a Neural Radiance Fields (NeRF) task (\cref{fig:1:gauss-nerf}), a Gaussian-based model trained with ESGD achieves lower reconstruction error (\cref{fig:1:nerf-gauss-esgd}, $0.0695$) compared to Adam (\cref{fig:1:nerf-gauss-adam}, $0.0743$), as seen in the error map (darker areas indicate lower error; zoom in $4\times$ for clarity).
}
\label{fig:main}
\end{center}
}]
\maketitle
\footnotetext[1]{$^*$Equal contribution. Correspondence to: Shin-Fang Chng $<$shinfangchng@gmail.com$>$, Hemanth Saratchandran $<$hemanth.saratchandran.adelaide.edu.au$>$.}

\begin{abstract}
Neural fields encode continuous multidimensional signals as neural networks, enabling diverse applications in computer vision, robotics, and geometry. While Adam is effective for stochastic optimization, it often requires long training times. To address this, we explore alternative optimization techniques to accelerate training without sacrificing accuracy. Traditional second-order methods like L-BFGS are unsuitable for stochastic settings. We propose a theoretical framework for training neural fields with curvature-aware diagonal preconditioners, demonstrating their effectiveness across tasks such as image reconstruction, shape modeling, and Neural Radiance Fields (NeRF)~\footnote{Source code is available at \url{https://github.com/sfchng/preconditioner_neural_fields.git}.}.
\end{abstract}    

\section{Introduction}

In recent years, neural fields, also known as implicit neural representations, have garnered significant attention in machine learning and computer graphics \cite{skorokhodov2021adversarial, sun2021coil, chen2021learning, li20223d, chen2022fully, sitzmann2019scene, nerf, siren}. Unlike traditional grid-based representations \cite{fridovich2022plenoxels, trading_pos, chen2022tensorf, sun2022direct}, neural fields implicitly encode shapes and objects through  neural networks. These representations are highly efficient for capturing complex, high-resolution signals, making them well-suited for tasks such as image synthesis \cite{skorokhodov2021adversarial, chen2021learning}, shape modeling \cite{nerf, sitzmann2019scene}, and robotics \cite{urain2022learning, chen2022neural}.

Classically, neural fields trained with ReLU activations suffer from spectral bias and limited fidelity \cite{rahaman2019spectral}. To address this, positional encoding (PE) layers \cite{tancik2020fourier} were introduced but at the cost of increased model complexity and noisy gradients. More recent work has explored non-standard activations, such as sine \cite{siren}, Gaussian \cite{ramasinghe22} and wavelets \cite{saragadam2023wire} which enable high-frequency encoding without positional encodings and produce well-behaved gradients \cite{saratchandran2023}.

While neural fields have enabled realistic and efficient representations of continuous signals, progress on minimizing the training time of such representations remains a significant research challenge. Currently, the Adam optimizer is the de facto algorithm for training neural fields, particularly in applications involving datasets where a stochastic training regime is essential. Yet, the preference for Adam over simpler alternatives such as stochastic gradient descent (SGD) has not been rigorously justified for neural fields, despite its widespread adoption. Notably, \cite{saratchandran2023} demonstrated substantial gains in training speed for neural fields using second-order optimizers such as L-BFGS \cite{nocedal1999numerical}. However, second-order methods are often impractical for stochastic settings \cite{saratchandran2023}, limiting their effectiveness for large-scale data, where stochastic training is essential. 


In this paper, we approach neural field training from the perspective of preconditioning. Preconditioners improve convergence by reducing the condition number of the optimization problem \cite{saad2003iterative, nocedal1999numerical, bradley2010algorithms}. We develop a theoretical framework linking Adam's performance to its implicit use of a diagonal preconditioner derived from the Gauss-Newton matrix and proceed to identify conditions under which curvature-aware diagonal preconditioners, such as those in ESGD (equilibrated stochastic gradient descent) \cite{dauphin2015equilibrated}, enhance training for neural fields with sine, Gaussian, or wavelet activations. For ReLU activations, we show that such preconditioning provides limited benefit.


Our experiments validate the proposed framework across a range of tasks, including 2D image reconstruction, 3D binary occupancy, and Neural Radiance Fields (NeRF). As demonstrated in \cref{fig:main}, ESGD consistently outperforms Adam for neural fields with wavelet, sine, and Gaussian activations, which, as predicted by our theory, is due to their better-conditioned Hessians. In contrast, Adam performs better for ReLU-PE activations, which are associated with poorly conditioned Hessians. Furthermore, we evaluated other curvature-aware algorithms, including ESGD, AdaHessian \cite{yao2021adahessian}, Shampoo \cite{gupta2018shampoo}, and K-FAC \cite{martens2015optimizing}, offering valuable insights into their effectiveness in stochastic training settings.


Our main contributions are:
\begin{enumerate}
    \item We present a theoretical framework for preconditioning in neural field training, uncovering key conditions and highlighting when preconditioning is beneficial.
    \item We empirically validate our framework, demonstrating the superiority of curvature-aware algorithms like ESGD over Adam across diverse tasks, including image reconstruction, 3D shape modeling, and NeRF.
\end{enumerate}

\section{Related Work}

\paragraph{Neural Fields} represent signals as weights within a neural network, using low-dimensional coordinates to maintain smoothness in the output \cite{trading_pos}. The influential work by Mildenhall et al. \cite{nerf} showcased the potential of neural fields for novel view synthesis, demonstrating their impressive generalization capabilities. Since then, neural fields have seen widespread application across image view synthesis, shape reconstruction, and robotics \cite{chen2019learning, Deng20, Genova_2020_CVPR, mulayoff2021implicit, Park_2019_CVPR, park2021nerfies, pumarola2021d, Rebain_2021_CVPR, Saito_2019_ICCV, siren, chng2022garf, wang2021nerf, Yu_2021_CVPR, saragadam2022miner, zhu2022nice, chen2023factor}. Traditionally, these approaches integrate a positional embedding layer to counteract the spectral bias introduced by ReLU activations \cite{rahaman2019spectral}. However, recent work has explored alternative, non-standard activations such as sine \cite{siren}, Gaussian \cite{ramasinghe22}, wavelets \cite{saragadam2023wire}, and sinc \cite{saratchandran2024sampling}. These activations have shown to outperform conventional approaches in various vision tasks, delivering enhanced performance in capturing high-frequency details and reducing reliance on positional embeddings \cite{siren, chng2022garf, ramasinghe22, saragadam2023wire, xia2022sinerf, lomas2022synthesis}.

\paragraph{Preconditioners} were initially introduced in numerical linear algebra \cite{saad2003iterative, greenbaum1997iterative} to transform ill-conditioned linear systems into well-conditioned forms, thereby accelerating the convergence of iterative solvers. Over time, preconditioning techniques have become integral to machine learning, where they enhance the efficiency and performance of optimization algorithms. Early applications in machine learning can be traced to foundational optimization methods, such as the conjugate gradient method and its variants \cite{nocedal1999numerical}. As machine learning problems have grown more complex—often involving sparse matrices, non-convex objectives, and large-scale datasets—researchers have developed specialized preconditioning strategies to meet these unique challenges \cite{bottou2018optimization, Zhaonan-Qu, recht2011hogwild, li2017preconditioned}. Many widely-used optimizers incorporate preconditioning, including methods like ADMM \cite{lin2016iteration}, mirror descent \cite{beck2003mirror}, quasi-Newton methods \cite{broyden1970convergence, goldfarb1970family, nocedal1999numerical}, and adaptive optimizers \cite{dauphin2015equilibrated, yao2021adahessian, duchi2011adaptive, gupta2018shampoo, martens2015optimizing}. These preconditioned optimizers have become essential for addressing the complex landscapes of modern machine learning models, driving improvements in both convergence speed and solution quality.

\section{Notation}\label{sec:notation}

The neural fields we consider in this paper will be defined by a depth $L$ neural network with layer widths 
$\{n_1,\ldots,n_L\}$. We let $X \in \R^{N\times n_0}$ denote the training data, with $n_0$ being the dimension of the input and $N$ being the number of training samples. The output at layer $k$ will be denoted by $F_k$ and is defined by
\begin{equation}\label{defn_net}
    F_k =  
    \begin{cases}
        F_{L-1}W_L + b_L, & k = L \\
        \phi(F_{k-1}W_k + b_k), & k \in [L-1] \\
        X, & k = 0
    \end{cases}
\end{equation}
where the weights $W_k \in \R^{n_{k-1}\times n_k}$ and the biases 
$b_k \in \R^{n_k}$ and $\phi$ is an activation applied component wise.
The notation
$[m]$ is defined by $[m] = \{1,\ldots,m\}$. We will denote the loss function that we train the neural fields with by
$\mathcal{L}$. In many experiments in the paper this will be the MSE loss \cite{prince2023understanding}.
Our theoretical results will be primarily for the case where $\phi$ is given by one of the activation 
sine, Gaussian, wavelet or ReLU. For more details on how these activations are used within the context of neural fields we refer the reader to \cite{siren, ramasinghe22, saragadam2023wire}. 

\section{Theoretical Framework}

\subsection{Overview of Preconditioners}\label{subsec:preconditioners}

Preconditioning enhances optimization convergence and efficiency by regulating 
curvature, particularly in stochastic gradient descent (SGD) 
\cite{dauphin2015equilibrated, Zhaonan-Qu}. High curvature directions in SGD require 
a small learning rate to avoid overshooting, leading to slow progress in low 
curvature directions. 

For general objective functions $f$ on parameters $\textbf{x} \in \R^n$. Preconditioning is done by employing a linear change of parameters 
$\widetilde{\textbf{x}} = D^{\frac{1}{2}}\textbf{x}$, where in practise 
$D^{\frac{1}{2}}$ is a non-singular matrix. With this change of parameters a new objective function $\widetilde{f}$, defined on 
$\widetilde{\textbf{x}}$, can be defined as
\begin{equation}\label{precond_obj_eqn}
    \widetilde{f}(\widetilde{\textbf{x}}) = f(\textbf{x}).
\end{equation}
Denoting the gradient of the original objective function by 
$\nabla f$ and its Hessian by $H(f) = \nabla^2 f(\textbf{x})$. The chain rule gives
\begin{align}
    \nabla \widetilde{f}(\widetilde{\textbf{x}}) &= D^{-\frac{1}{2}}\nabla f 
    \label{precond_grad_eqn} \\
    H(\widetilde{f})(\widetilde{\textbf{x}}) &= 
    (D^{-\frac{1}{2}})^T (H(f)(\textbf{x})) (D^{-\frac{1}{2}}).
      \label{precond_hess_eqn}
   \end{align}

Equation \eqref{precond_grad_eqn} exhibits the key idea of training with a 
preconditioner. If we consider gradient descent on the transformed parameter space, 
a 
gradient descent iteration takes the form 
\begin{equation}\label{grad_update_transformed}
    \widetilde{\textbf{x}}_t = \widetilde{\textbf{x}}_{t-1} - 
\eta\nabla_{\widetilde{\textbf{x}}_{t-1}} \widetilde{f}
\end{equation}
Applying equations
\eqref{precond_obj_eqn} and
\eqref{precond_grad_eqn} this corresponds to doing the update
\begin{equation}\label{grad_update_precond}
    \textbf{x}_t = \textbf{x}_{t-1} - 
\eta D^{-\frac{1}{2}}\nabla_{\textbf{x}_{t-1}}f.
\end{equation}
The main endeavour of preconditioning for optimization is to find a 
preconditioning matrix $D$ such that the new Hessian, given by equation
\eqref{precond_hess_eqn}, has equal curvature in all directions. This will 
allow the update \eqref{grad_update_precond} to move faster along small 
curvature directions, leading to faster convergence. 
Algorithm \ref{Alg:precond_sgd} gives the basic pseudocode for preconditioned SGD.

\begin{algorithm}
\caption{Preconditioned SGD}\label{Alg:precond_sgd}
\begin{algorithmic}
\Require \text{initial point } $x_0$, iterations $N$, learning rate $\eta$, preconditioner $D$.
\For{$t = 0,\cdots, N$} 
\State $g_t \gets \nabla_{x_t} f$ 
\State $D_t \gets D(x_t)$ 
\State $x_{t+1} \gets x_t -  \eta D(x_t)^{-\frac{1}{2}}g_t$ 
\EndFor
\end{algorithmic}
\end{algorithm}

From \cref{precond_hess_eqn} we see that the best preconditioner to 
use would be the absolute value of the Hessian itself. In the case of a positive 
definite Hessian, this is precisely what Newton's method does 
\cite{nocedal1999numerical}. The main issue with this preconditioner is that the cost of storing and inverting the preconditioner will be $\mathcal{O}(n^3)$ for an objective function of $n$ parameters \cite{nocedal1999numerical}. This will be extremely costly for the training of overparameterized neural networks whose Hessians are extremely large.

We list some of the main preconditioners that are being used in the literature. 
\paragraph{Diagonal Preconditioners:}\label{para:diag_precond} are matrices that represent some form of a 
diagonal approximation of the Hessian. Examples include the Jacobi preconditioner, 
which is given by $D^J = |diag(H)|$, where $|\cdot|$ is element-wise absolute value and 
$H$ is the Hessian. In practical settings, the absolute value is taken so that the preconditioner is positive semidefinite.
The equilibration preconditioner is a diagonal matrix which has the 
2-norm of each row (or column) on the diagonal: 
$D^E = \vert\vert H_{i,\cdot}\vert\vert$. 
The preconditioners $D^J$ and $D^E$ are known as curvature-aware preconditioners as they seek to give a good approximation to the Hessian.
Diagonal preconditioners are often used because of their ability to be stored and inverted in linear time. Diagonal preconditioners are also known as adapative learning rates in the literature.
The algorithm Equilibrated Stochastic Gradient Descent 
(ESGD) \cite{dauphin2015equilibrated} employs $D^E$ as a preconditioner to condition stochastic 
gradient descent. In \cite{dauphin2015equilibrated}, it is shown that applying ESGD leads to faster training times
for autoencoder tasks when compared to vanilla SGD, RMSProp \cite{ruder2016overview} and a preconditioned SGD using $D^J$ as preconditioner. 
AdaHessian \cite{yao2021adahessian} is a recent algorithm that applies a diagonal
preconditioner in the setting of moving averages and thus can be seen as a way of 
enhancing the Adam optimizer \cite{kingma2014adam}. 

\paragraph{Kronecker Factored Preconditioners:} are preconditioners of the form
$D = A^TA$, where $A = A_1 \otimes \cdots \otimes A_k$ and $\otimes$ denotes the 
Kronecker product.
The algorithm K-FAC \cite{martens2015optimizing} employs a Kronecker factored 
preconditioner.

\textbf{Others:} There are several second order optimisation algorithms that use 
preconditioners that do not fall into the above three categories such as, The Gauss-Newton method, The Conjugate Gradient method, Limited memory Broyden-Fisher-Golfarb-
Shanno (L-BFGS) \cite{nocedal1999numerical}. It was shown in \cite{saratchandran2023} that these algorithms do not train well in stochastic settings. 

\subsection{Condition Number}\label{subsec:condition number}

In \cref{subsec:preconditioners}, we explained that SGD will move faster 
through the loss landscape when the curvatures in each direction are similar. 
The regularity of the loss landscape at a point is quantified via the condition number of the Hessian.
\begin{definition}
    Given a matrix $A$ we define the condition number $\kappa$ of A by
    \begin{equation}\label{cond_number_defn}
        \kappa(A) := \frac{\sigma_{\max}(A)}{\sigma_{\min}(A)},
    \end{equation}
\end{definition}
where $\sigma$ is the eigenvalue.
Using \cref{precond_hess_eqn} we observe that if we take a preconditioner $D$ such that
$\kappa((D^{-\frac{1}{2}})^T (H(f)(\textbf{x}))$ $ (D^{-\frac{1}{2}}))$ is closer to 1 
than $\kappa(H)$, then this implies that the preconditioner $D$ is regularising the 
curvatures of the landscape to be equal in each direction, which yields a better training landscape for a gradient descent based algorithm \cite{nocedal1999numerical}.

\begin{figure}[t]
    \centering
    \begin{subfigure}{0.49\linewidth} \includegraphics[width=\linewidth]{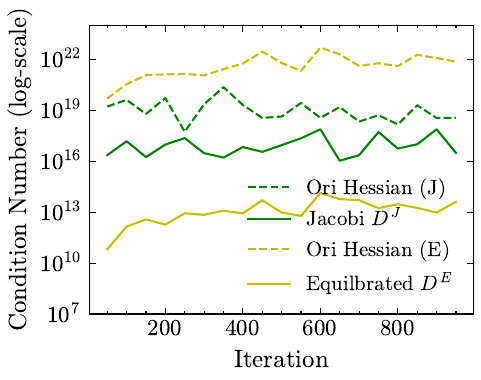}\caption{}\label{fig:theory_condition_number}\end{subfigure}
    \begin{subfigure}{0.49\linewidth} \includegraphics[width=\linewidth]{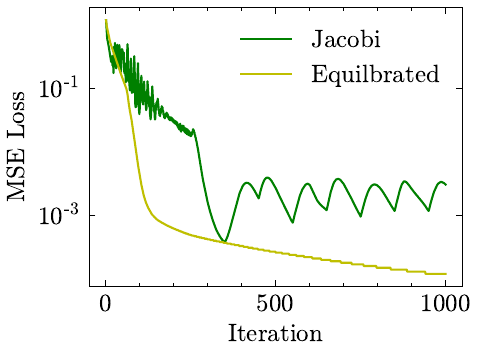}\caption{}\label{fig:theory_1d_convergence}\end{subfigure}
    \vspace{-1em}
    \caption{\cref {fig:theory_condition_number} compares the condition number (\ref{cond_number_defn}) on a 1D signal regression task \textit{before} (\textbf{dotted} line) and \textit{after} (\textbf{solid} line) applying equilbrated $D^E$ and Jacobi $D^J$ preconditioners. The equilibrated preconditioner significantly reduces the Hessian's condition number, as shown by the larger gap between the yellow solid and dotted lines compared to the green solid and dotted lines representing the Jacobi preconditioner, thus achieves faster convergence as seen in \cref{fig:theory_1d_convergence}. }
    \label{fig:condition_number}
\end{figure}

To empirically illustrate the concept of preconditioning, we examined a one-dimensional curve fitting scenario. Specifically, we considered a feedforward network with two hidden layers, each containing 16 neurons, tasked with fitting the signal $f(x) = \sin(2\pi x) + \sin(6\pi x)$ using a Gaussian activation \cite{ramasinghe22}. During optimization with gradient descent, we applied two types of preconditioners: an equilibrated preconditioner $D^E$ and a Jacobi preconditioner $D^J$. At each optimization step, we computed both the original Hessian associated with gradient descent and the transformed Hessian using \cref{precond_hess_eqn}. As shown in \cref{fig:condition_number}, the results demonstrate that applying either the Jacobi or equilibrated preconditioner consistently reduced the condition number of the Hessian compared to the unconditioned case, highlighting the effectiveness of preconditioning in improving optimization.


\subsection{Training with Adam}\label{subsec: Adam_INRs}

\begin{figure*}[t]
  \centering
\begin{subfigure}{0.33\linewidth} \includegraphics[width=\linewidth]{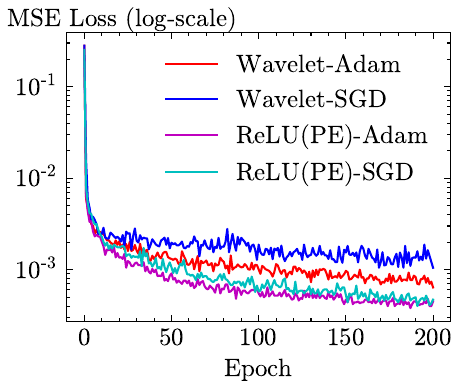}\caption{}\label{fig:theory_img_sgd_vs_adam}\end{subfigure}
\begin{subfigure}{0.32\linewidth} \includegraphics[width=\linewidth]{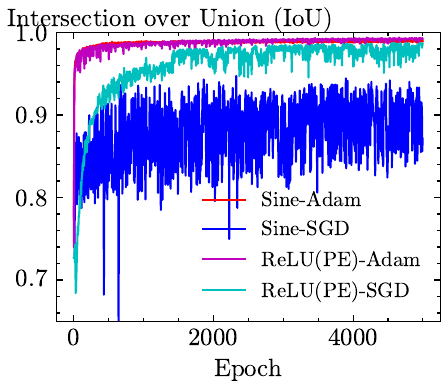}\caption{}\label{fig:theory_occupancy_sgd_vs_adam}\end{subfigure}
\begin{subfigure}{0.33\linewidth} \includegraphics[width=\linewidth]{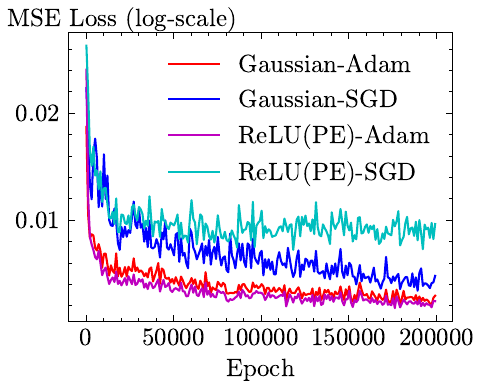}\caption{}\label{fig:theory_nerf_sgd_vs_adam}\end{subfigure}
\vspace{-1em}
\caption{Comparison of training convergence for neural fields with different activations -- ReLU with positional encoding (ReLU (PE)), wavelet, sine and Gaussian -- using Adam and SGD optimizers across three tasks: 2D image reconstruction (\cref{fig:theory_img_sgd_vs_adam}), 3D binary occupancy reconstruction (\cref{fig:theory_occupancy_sgd_vs_adam}), and NeRF (\cref{fig:theory_nerf_sgd_vs_adam}). In all cases, the Adam optimizer outperforms SGD, indicating that leveraging curvature information facilitates faster convergence.}
\label{fig:sgd_vs_adam}
\end{figure*}

In this section, we aim to explain why training with Adam is generally more efficient than training with SGD for a variety of neural fields. While this observation is well-known within the community, we include it here to provide context and motivation for the theoretical framework developed in \cref{sec:main_theorems}.

Given an objective function $L(\theta) = |f(\theta)|^2$ for a non-linear function 
$f$. The Jacobian $J_L$ of $L$ can be computed via the chain rule: $J_L = 2J_f^Tf$, 
where $J_f$ denotes the Jacobian of $f$. The Hessian of $L$, $H_L$, can be computed 
by an application of the chain rule
\begin{equation}\label{hessian_L}
    H_L = 2J_f^TJ_f + 2H_ff.
\end{equation}
We can then approximate the Hessian $H_L$ by the term $2J_f^TJ_f$. Leading to the first order approximation
\begin{equation}
    H_L \approx 2J_f^TJ_f 
         = \frac{1}{2|f|^2}J_L^TJ_L 
         = \frac{1}{2L}J_L^TJ_L. \label{hess_L_approx3}
\end{equation}
\cref{hess_L_approx3} yields the basic idea of the Gauss-Newton method. By taking 
the Gauss-Newton matrix $J_L^TJ_L$ as an approximation to the Hessian $H_L$, we can 
perform an optimization update that uses the step: 
$\Delta_{GN} = -\frac{1}{J^TJ}\nabla L$. The Gauss-Newton matrix is in general only 
positive semi-definite. Therefore, when implementing the algorithm one often adds a 
small damping factor $\lambda$ so the implemented step is 
$\Delta_{GN} = -\frac{1}{J^TJ + \lambda}\nabla L$. A similar derivation can be made for more general loss functions used in non-linear least squares type regression problems \cite{nocedal1999numerical}. It follows that
the Gauss-Newton method can be seen as a preconditioned SGD that 
uses a first order (first derivatives) Hessian approximate preconditioner given by the Gauss-Newton matrix $J^TJ$.

Using the above observations we now show how to view Adam as an optimizer that applies a diagonal preconditioner.
The Adam optimizer involves two key steps. The first is to compute a first moment 
estimate of the form $m_t = \beta_1m_{t-1} + (1-\beta_1)g_t$, where $g_t$ is the 
gradient of a given objective function $f$ at iteration $t$, and $m_0 = 0$. 
The second is to 
compute a second moment estimate 
$v_t = \beta_2v_{t-1} + (1-\beta_2)g_t^2$, where $g_t^2 = g_t \odot g_t$,  
$\odot$ denotes the Hadamard product, and $v_0 = 0$. The numbers 
$0 < \beta_1 < \beta_2 < 1$ are hyperparameters. The algorithm then applies a 
bias correction to the above first and second moment estimates, and then takes a moving average to perform an update step based on $-\frac{m_t}{\sqrt{v_t}}$,
see \cite{kingma2014adam} for 
details. 

For the following discussion, 
we will only focus on analyzing the second raw moment estimate, as this is 
related to the Gauss-Newton matrix. 
Given an objective function $f$, observe that we can express $g_t^2 = g_t \odot g_t = 
Diag(J^TJ)$, where $J$ denotes the Jacobian of $f$. Thus the second raw moment 
estimate of Adam can be recast as: 
\begin{equation}\label{second_moment_Adam_via_GN}
    v_t = \beta_2v_{t-1} + (1-\beta_2)Diag(J_t^TJ_t)
\end{equation}
where $J_t$ denotes the Jacobian at iteration $t$. 
Thus we see that Adam 
applies a preconditioner given by the diagonal of the Gauss-Newton matrix. Since the 
Gauss-Newton matrix can be seen as a first order approximation to the Hessian, Adam 
can be seen as using an approximation to curvature to adapt the learning rate for faster 
optimization. This suggests that Adam can outperform SGD which employs no 
preconditioner. 

We empirically validated this insight across three distinct neural field tasks: image reconstruction, binary occupancy, and NeRF. For each task, we compared the performance of ReLU(PE) activated networks trained with both SGD and Adam against networks using wavelet, sine, and Gaussian activations. As shown in \cref{fig:sgd_vs_adam}, Adam consistently outperforms SGD in all cases. This suggests that Adam's use of a preconditioner significantly enhances its performance for neural field applications, see \cref{sec:experiments} for the details of the architectures and experimental setup.

\subsection{Main theorems}\label{sec:main_theorems}

\begin{figure}[t]
    \centering
    \includegraphics[width=0.75\linewidth]
    {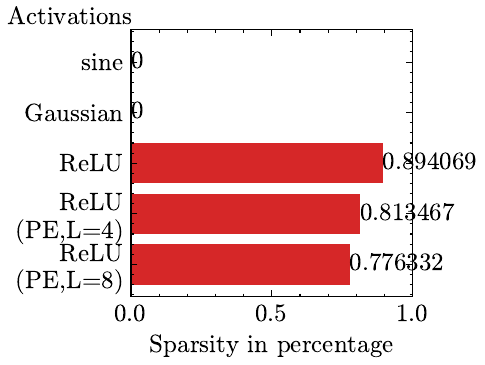}
    \vspace{-1em}
    \caption{Percentage sparsity of Hessian-vector product matrices across \textbf{all} layers \textbf{throughout the training} process of a 2D image reconstruction task, using ESGD to train networks with different activations. The x-axis ranges from $0$ to $1$, with values closer to 1 indicating higher sparsity. }
    \label{fig:sparsity1}
\end{figure}

In \cref{subsec: Adam_INRs}, we discussed how Adam can be viewed as applying a diagonal preconditioner based on the diagonal of the Gauss-Newton matrix, which uses first derivatives to approximate the Hessian matrix. Here, we aim to identify the architectural properties that enable effective training of neural fields with diagonal preconditioners that give a better approximation to the Hessian. Our primary theorems demonstrate that the choice of activation function is crucial for enabling efficient training with such preconditioners, highlighting its central role in achieving effective curvature-aware optimization for neural fields.

Recall \cref{subsec:preconditioners} that diagonal preconditoners such as the equilibrated precondition $D^E$ or the Jacobi preconditioner $D^J$ all involve the Hessian matrix. In the context of neural networks the Hessian of the loss landscape is often an extremely large matrix. Thus in practical settings one often replaces the Hessian with a Hessian vector product which is much cheaper to store 
\cite{nocedal1999numerical}.

Our first theorem shows that the Hessian vector product of the loss landscape associated to a 
sine, Gaussian, wavelet or sinc activated neural field produces a dense vector. 

\begin{theorem}\label{thm:hess_vec_gauss/sine}
Let $F$ denote a sine, Gaussian, wavelet or sinc activated neural field. 
Let $\mathcal{L}(\theta)$ denote the MSE loss associated to $F$ and a training 
set $(X, Y)$. Let $H$ denote the Hessian of $\mathcal{L}$ at a fixed parameter 
point $\theta$. Then given a non-zero vector $v$ the Hessian vector product $Hv$ 
will be a dense vector. 
\end{theorem}

While it may seem rather costly for a sine- or Gaussian-activated INR to have dense 
Hessian vector products when applied to the MSE loss function, we point out that the 
cost is much less than storing a dense Hessian.

\begin{theorem}\label{thm:hess_vec_relu}
Let $F$ denote a ReLU/ReLU(PE)-activated neural field. 
Let $\mathcal{L}(\theta)$ denote the MSE loss associated to $F$ and a training 
set $(X, Y)$. Let $H$ denote the Hessian of $\mathcal{L}$ at a fixed parameter 
point $\theta$. Then given a non-zero vector $v$ the Hessian vector product $Hv$ 
will be a sparse vector. 
\end{theorem}

\cref{thm:hess_vec_gauss/sine} and \cref{thm:hess_vec_relu} also hold for the binary cross-entropy loss. The proofs of \cref{thm:hess_vec_gauss/sine} and \cref{thm:hess_vec_relu} are given in the supp.

\cref{thm:hess_vec_relu} implies that when applying a 
preconditioner, built from a Hessian vector 
product, to a ReLU/ReLU(PE) INR, such as 
ESGD \cite{dauphin2015equilibrated} or AdaHessian \cite{yao2021adahessian}, only a few components of the gradient will be 
scaled during the update step. This means that the preconditioner will not be adding 
much to the optimization process. On the other hand, \cref{thm:hess_vec_gauss/sine} implies that applying a preconditioner built from the Hessian to 
the 
gradient of the MSE loss of a sine, Gaussian and wavelet-activated neural field should help optimization 
as many of the gradient components will get scaled by a quantity depending on the 
curvature of the loss landscape. This yields a clear difference between ReLU/ReLU(PE) activated neural fields and those activated with a sine, Gaussian, wavelet activation. In particular, \cref{thm:hess_vec_gauss/sine} implies that neural fields admitting a sine, Gaussian, wavelet activation have the ability to be trained efficiently with a diagonal curvature aware preconditioner such as the equilibrated preconditioner $D^E$ defined in \cref{subsec:preconditioners}.

\cref{fig:sparsity1} 
shows the results of a Hessian vector product of the MSE loss of a ReLU(PE) and Gaussian neural field for an image reconstruction task for the weights in all layers of the network
layer throughout training.
The random vector was chosen from a Rademacher distribution and the reported figures are given as the mean of the Hessian of all layers averaged over all iterations of training.
As demonstrated by the experiment, the ReLU(PE) network generates a sparse Hessian-vector product, as predicted by \cref{thm:hess_vec_relu}, while the Gaussian network produces a dense vector, as indicated by \cref{thm:hess_vec_gauss/sine}.
Furthermore, as we will observe in \cref{sec:experiments}, this sparsity helps explain why Adam performs well for neural fields with ReLU (ReLU(PE)) activations, whereas an algorithm like ESGD \cite{dauphin2015equilibrated} does not. This is because Adam does not rely on Hessian-vector product.


\section{Experiments}\label{sec:experiments}
\begin{figure*}[t]
  \centering
\begin{subfigure}{0.35\linewidth} \includegraphics[width=\linewidth]{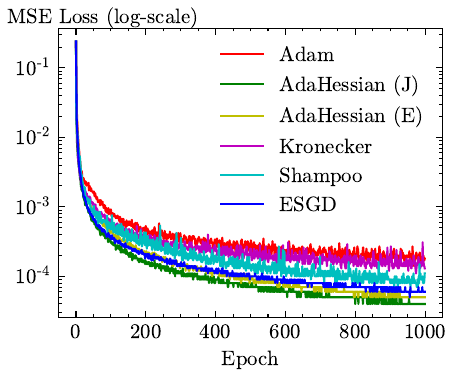}\caption{}\label{fig:exp_img_gauss_convergence}\end{subfigure}
\begin{subfigure}{0.35\linewidth} \includegraphics[width=\linewidth]{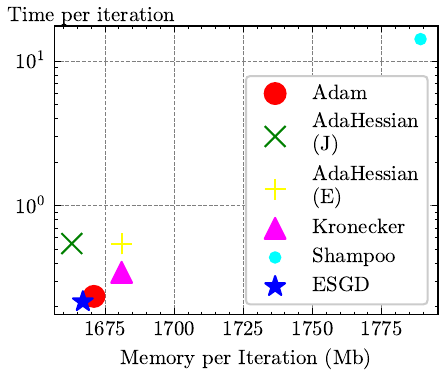}\caption{}\label{fig:exp_img_gauss_computation}\end{subfigure}
\vspace{-1em}
\caption{\textbf{Comparison of training convergence and computational complexity for various preconditioners.} We evaluated a Gaussian-activated neural field on the \textit{lion} instance from the DIV2K dataset. ESGD demonstrates superior convergence compared to other preconditioners, striking a balance between accuracy and computational efficiency. \textbf{Note}: We provide the time-based comparisons in the supp. (Sec 3.1). Similar analysis for other activations and additional instances from the DIV2K dataset are available in the supp. (Sec. 3.2).}\label{fig:exp_img_gauss_convergence_and_computation}
\end{figure*}
We compared various preconditioners over popular benchmarks~\cite{hertz2021sape,siren,saratchandran2023,lindell2022bacon,shabanov2024banf}: 2D image reconstruction, 3D shape reconstruction and novel view synthesis with NeRF. Our primary focus was on curvature-aware preconditoners that are well-suited for the stochastic training of neural network. Specifically, we compared the following methods: ESGD~\cite{dauphin2015equilibrated}\footnote{https://github.com/crowsonkb/esgd}, AdaHessian~\cite{yao2021adahessian}\footnote{https://github.com/amirgholami/adahessian}, Shampoo~\cite{gupta2018shampoo}\footnote{https://github.com/google-research/google-research/tree/master/scalable\_shampoo/pytorch}, and Kronecker-based preconditioner~\cite{martens2015optimizing}\footnote{https://github.com/Thrandis/EKFAC-pytorch}. We provide the details of each algorithm including the reproducibility details of each experiment and machines we used to run each experiment in the supp. In the main paper, we focused on analyzing Gaussian networks, see supp. for additional analysis for other non-traditional activations and other neural field tasks.
\vspace{-1em}
\paragraph{Remark for AdaHessian} We observed that in the implementation of AdaHessian, an equilbrated preconditioner $D^{E}$ is utilized, even though the paper primarily focuses on using a Jacobi preconditioner~\cite{dauphin2015equilibrated}. Therefore, we introduced a Jacobi preconditioner $D^{J}$ into AdaHessian, which we refer to as \textbf{AdaHessian(J)}. We use the term \textbf{AdaHessian(E)} for the original implementation. 
\vspace{-1em}
\paragraph{Remark for ESGD} We further modified the ESGD by applying the squared Hessian diagonal preconditioner in two settings: using an exponential moving average (ESGD) and using an inifinity norm (ESGD-max)~\cite{kingma2014adam}. We also opted to recompute the $D^{E}$ for every $N$ iterations for computational efficiency, we use $N=100$ for all the experiments.

\subsection{Image Reconstruction}
We evaluated on a 2D image reconstruction task, where given pixel coordinates $\mathbf{x} \in \mathbb{R}^2$, we optimized a neural network to regress the corresponding RGB values $c \in \mathbb{R}^3$. We used a 5-layer Gaussian-activated network with 256 neurons. We used $\sigma=0.05$ for the Gaussian activation. We trained the network using mini-batches of size $512$.
We compared the training convergence of different preconditioners and the Adam optimizer on the \textit{lion} instance from the DIV2K dataset~\cite{Agustsson_2017_CVPR_Workshops} in \cref{fig:exp_img_gauss_convergence}. Additional results on the rest of the DIV2K dataset are provided in the supp. As predicted by \cref{thm:hess_vec_gauss/sine}, Gaussian network equipped with the dense Hessian-vector product exhibits faster convergence when optimized with curvature-aware preconditioned methods compared to Adam. Notably, ESGD demonstrates faster convergence and has comparable training time and memory usage per iteration compared to Adam. 

ESGD and AdaHessian obtain Hessian information by computing Hessian-vector products, which makes them very computational efficient. Each Hessian-vector product is generally as fast as a single gradient computation, as demonstrated in \cref{fig:exp_img_gauss_computation}. In contrast, other preconditioners, particularly Shampoo are more computationally expensive because they involve Kronecker product computations for preconditioner matrices, see supp. for details.
\begin{table*}[t]
    \centering
    \begin{tabular}{ll|ccccc|ccccc}
    \toprule
         \multicolumn{2}{l|}{} &\multicolumn{2}{c}{} & \multicolumn{3}{c|}{\textbf{Adam}}  &\multicolumn{2}{c}{} &\multicolumn{3}{c}{\textbf{ESGD}}     \\
        \multicolumn{1}{l}{Scene} & \multicolumn{1}{c|}{Train}  
         & \multicolumn{1}{c}{Iteration } & \multicolumn{1}{c}{Time} 
         & \multicolumn{3}{c|}{Test}& \multicolumn{1}{c}{Iteration} & \multicolumn{1}{c}{Time }  & \multicolumn{3}{c}{Test} \\ \cline{5-7} \cline{10-12}
         \multicolumn{1}{l}{} & \multicolumn{1}{c|}{PSNR}  
         & \multicolumn{1}{c}{$\downarrow$} & \multicolumn{1}{c}{($s$)$\downarrow$} 
         & \multicolumn{1}{c}{PSNR $\uparrow$} & \multicolumn{1}{c}{SSIM $\uparrow$} & \multicolumn{1}{c|}{LPIPS $\downarrow$ } & \multicolumn{1}{c}{$\downarrow$} & \multicolumn{1}{c}{($s$)$\downarrow$}
         & \multicolumn{1}{c}{PSNR $\uparrow$} & \multicolumn{1}{c}{SSIM $\uparrow$} & \multicolumn{1}{c}{LPIPS $\downarrow$ } \\
    \midrule
         \multicolumn{1}{c}{fern} & \multicolumn{1}{c|}{25.38}  
         & \multicolumn{1}{c}{200K} & \multicolumn{1}{c}{368.43} 
         & \multicolumn{1}{c}{24.38} & \multicolumn{1}{c}{0.74} & \multicolumn{1}{c|}{0.28} & \multicolumn{1}{c}{120K} & \multicolumn{1}{c}{\textbf{270.81}} 
         & \multicolumn{1}{c}{24.41} & \multicolumn{1}{c}{0.74} & \multicolumn{1}{c}{0.30} \\
         \multicolumn{1}{c}{flower} & \multicolumn{1}{c|}{29.17}  
         & \multicolumn{1}{c}{200K} & \multicolumn{1}{c}{469.31} 
         & \multicolumn{1}{c}{25.67} & \multicolumn{1}{c}{0.78} & \multicolumn{1}{c|}{0.12} & \multicolumn{1}{c}{120K} & \multicolumn{1}{c}{\textbf{263.93}} 
         & \multicolumn{1}{c}{25.65} & \multicolumn{1}{c}{0.78} & \multicolumn{1}{c}{0.13} \\
         \multicolumn{1}{c}{horns} & \multicolumn{1}{c|}{24.95}  
         & \multicolumn{1}{c}{60K} & \multicolumn{1}{c}{\textbf{130.82}} 
         & \multicolumn{1}{c}{21.82} & \multicolumn{1}{c}{0.70} & \multicolumn{1}{c|}{0.34} & \multicolumn{1}{c}{120K} & \multicolumn{1}{c}{227.58} 
         & \multicolumn{1}{c}{20.78} & \multicolumn{1}{c}{0.63} & \multicolumn{1}{c}{0.44} \\
         \multicolumn{1}{c}{orchids} & \multicolumn{1}{c|}{23.3}  
         & \multicolumn{1}{c}{200K} & \multicolumn{1}{c}{501.90} 
         & \multicolumn{1}{c}{19.56} & \multicolumn{1}{c}{0.61} & \multicolumn{1}{c|}{0.24} & \multicolumn{1}{c}{120K} & \multicolumn{1}{c}{\textbf{269.01}} 
         & \multicolumn{1}{c}{19.74} & \multicolumn{1}{c}{0.61} & \multicolumn{1}{c}{0.21} \\
         \multicolumn{1}{c}{room} & \multicolumn{1}{c|}{31.20}  
         & \multicolumn{1}{c}{110K} & \multicolumn{1}{c}{\textbf{194.05}} 
         & \multicolumn{1}{c}{31.60} & \multicolumn{1}{c}{0.93} & \multicolumn{1}{c|}{0.12} & \multicolumn{1}{c}{140K} & \multicolumn{1}{c}{201.88} 
         & \multicolumn{1}{c}{30.52} & \multicolumn{1}{c}{0.90} & \multicolumn{1}{c}{0.20} \\
         \multicolumn{1}{c}{fortress} & \multicolumn{1}{c|}{30.18}  
         & \multicolumn{1}{c}{140K} & \multicolumn{1}{c}{\textbf{333.70}} 
         & \multicolumn{1}{c}{28.93} & \multicolumn{1}{c}{0.81} & \multicolumn{1}{c|}{0.15} & \multicolumn{1}{c}{140K} & \multicolumn{1}{c}{334.56} 
         & \multicolumn{1}{c}{28.6} & \multicolumn{1}{c}{0.80} & \multicolumn{1}{c}{0.17} \\ 
         \multicolumn{1}{c}{leaves} & \multicolumn{1}{c|}{21.36}  
         & \multicolumn{1}{c}{200K} & \multicolumn{1}{c}{482.34} 
         & \multicolumn{1}{c}{19.56} & \multicolumn{1}{c}{0.61} & \multicolumn{1}{c|}{0.24} & \multicolumn{1}{c}{120K} & \multicolumn{1}{c}{\textbf{271.25}} 
         & \multicolumn{1}{c}{19.43} & \multicolumn{1}{c}{0.59} & \multicolumn{1}{c}{0.25} \\
         \multicolumn{1}{c}{trex} & \multicolumn{1}{c|}{25.50}  
         & \multicolumn{1}{c}{80K} & \multicolumn{1}{c}{\textbf{164.96}} 
         & \multicolumn{1}{c}{22.04} & \multicolumn{1}{c}{0.71} & \multicolumn{1}{c|}{0.31} & \multicolumn{1}{c}{120K} & \multicolumn{1}{c}{209.25} 
         & \multicolumn{1}{c}{22.21} & \multicolumn{1}{c}{0.75} & \multicolumn{1}{c}{0.23} \\
   \bottomrule
    \end{tabular}
    \caption{\textbf{Quantitative results of NeRF on all instances from the LLFF dataset}~\cite{mildenhall2019local}. The reported time solely refers to the optimization update step. \textbf{Note:} Results for the Blender synthetic~\cite{nerf} and BLEFF~\cite{wang2021nerf} datasets are available in supp.}\label{exp:nerf}
    \label{tab:nerf}
\end{table*}

\subsection{3D Binary Occupancy Reconstruction}
We evaluated on learning a binary occupancy field, where given 3D point coordinates $\mathbf{x} \in \mathbb{R}^{3}$, we optimized a neural network to represent the occupancy field of 3D shapes. Following existing works~\cite{hertz2021sape}, we sampled one million 3D points for training -- one-third of the points were sampled uniformly within the volume, and the remaining two-thirds of the points were sampled near the mesh surface and perturbed with random Gaussian noise using sigma of 0.1 and 0.01, respectively~\cite{hertz2021sape}. We trained the network using binary cross-entropy (BCE) loss, which compares the predicted occupancy and the true occupancy. We used a 5-layer Gaussian network with 256 neurons and $\sigma=0.09$. We trained the network using mini-batches of size $50000$.

\begin{figure}[b]
    \centering  
    \includegraphics[width=0.80\linewidth]{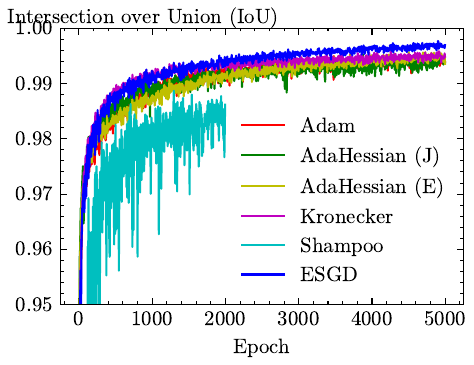}
    \vspace{-1em}
    \caption{\textbf{Comparison of training convergence for various preconditioners.} We evaluated a Gaussian-activated neural field on a 3D binary occupancy reconstruction task on the \textit{armadillo} instance. ESGD demonstrates superior convergence compared to other preconditioners, striking a balance between accuracy and computational efficiency. \textbf{Note}: Similar analysis for other activations are available in the supp.}\label{fig:exp_occupancy_gauss_convergence}
\end{figure}

\begin{figure}[t]
    \centering  
    \begin{subfigure}{0.49\linewidth} \includegraphics[width=\linewidth]{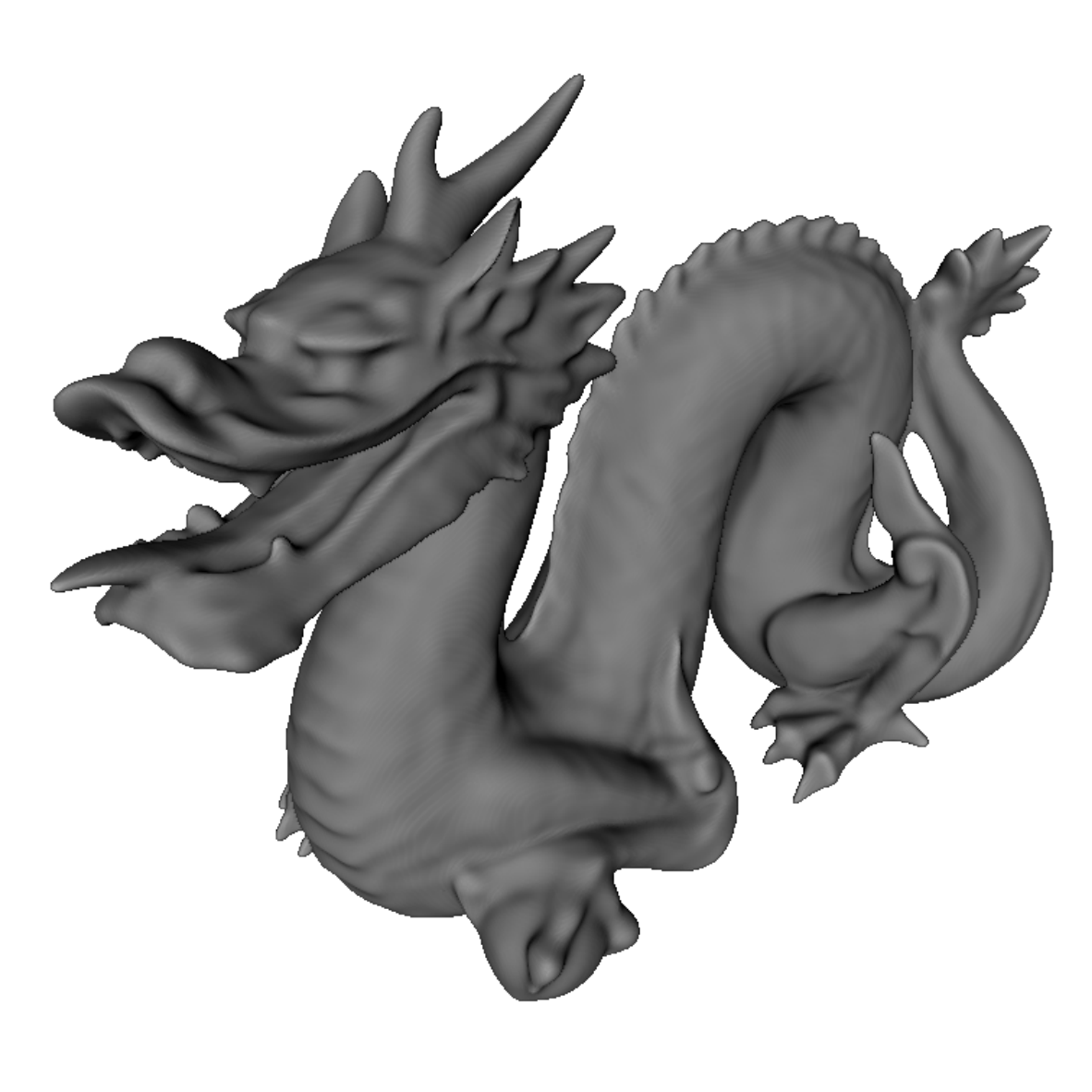}\caption{}\label{fig:exp_occupancy_dragon_adam}\end{subfigure}
    \begin{subfigure}{0.49\linewidth} \includegraphics[width=\linewidth]{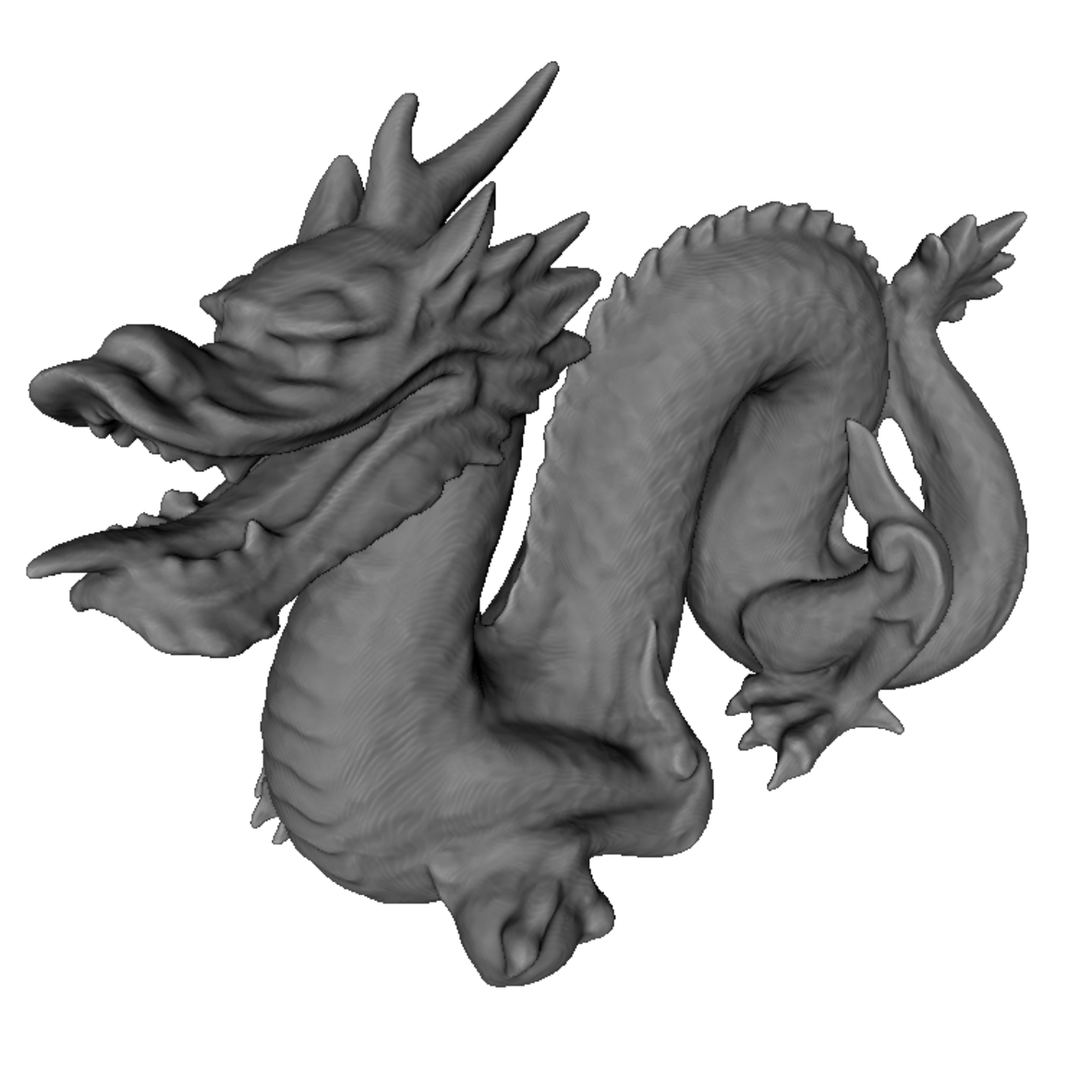}\caption{}\label{fig:exp_occupancy_dragon_esgd}\end{subfigure}
    \vspace{-1em}
    \caption{\textbf{Qualitative results for the \textit{dragon} instance from Stanford 3D Scanning dataset}. Compared to the Gaussian-based neural field trained with Adam (\cref{fig:exp_occupancy_dragon_adam}), ESGD (\cref{fig:exp_occupancy_dragon_esgd}) has reconstructed the shapes with significantly improved fidelity at epoch 500 (zoom in $4 \times$ for better visibility). \textbf{Note:} Qualitative results for the rest of the instances from Stanford 3D Scanning Dataset are available in supp.}\label{fig:occupancy_dragon}
\end{figure}

We compared the training convergence of different preconditioners and the Adam optimizer on the \textit{armadillo} instance from the Stanford 3D Scanning dataset.~\footnote{http://graphics.stanford.edu/data/3Dscanrep/} \cref{fig:exp_occupancy_gauss_convergence} shows that ESGD exhibits superior convergence than other preconditioner-type optimizers. We showcase a qualitative result comparing the mesh reconstruction of ESGD and Adam given the same number of iteration in \cref{fig:occupancy_dragon}. Interestingly, when dealing with higher mode signals, both AdaHessian and Shampoo no longer demonstrated superior convergence, unlike in the 2D image reconstruction task in~\cref{fig:exp_img_gauss_convergence}. We speculate that the local Hessians may become noisy with higher mode signals, thereby impacting the gradient updates. In contrast, this noise did not significantly affect ESGD's convergence because its preconditioner updates were performed every $N$ iterations. Hence, we will focus on comparing Adam and ESGD for the experiments in the following section.

\subsection{Neural Radiance Fields (NeRF)}\label{subsec:nerf}

Neural Radiance Fields (NeRF) have recently emerged as a pioneering approach in  novel view synthesis, utilizing neural networks to model volumetric fields of 3D scenes using multi-view 2D images~\cite{nerf,kilonerf,chng2022garf,lin2021barf,chng2024invertible}.
Specifically, given 3D points $\mathbf{x} \in \mathbb{R}^3$ and viewing direction, NeRF estimates the 5D radiance field of a 3D scene which maps each input 3D coordinate to its corresponding volume density $\sigma \in \mathbb{R}$ and directional emitted color $\mathbf{c} \in \mathbb{R}^3$~\cite{nerf,lin2021barf,chng2022garf}. We used an 8-layer Gaussian-activated network with 256 neurons and $\sigma=0.1$. Recall that in the previous section we demonstrated that ESGD exhibited superior performance in terms of convergence and computational complexity compared to other preconditioners. Therefore, we focused on comparing Adam and ESGD here. We determined the same training PSNR at which both model converged and reported the corresponding optimization metrics, including the number of iterations and the total time taken for the update computation. We evaluated the PSNR on a hold-out test scene. Table~\ref{exp:nerf}. shows that ESGD achieves on average faster convergence than Adam, see more results in the supp. 


\section{Limitations}
While our work theoretically established the suitability of curvature-aware preconditioners for training neural fields with non-traditional activations such as sine, Gaussian, and wavelets, it also highlighted their ineffectiveness for neural fields employing ReLU-PE activations. Specifically, we showed that Adam performs well for ReLU-PE due to its use of a diagonal preconditioner derived from the Gauss-Newton matrix, rather than relying on Hessian vector products. 
However, this raises an important limitation of our work: we did not explore whether alternative preconditioners, potentially more effective than the Gauss-Newton matrix, could be developed to improve the training of ReLU-PE activated neural fields beyond what Adam achieves. This is particularly relevant since ReLU-PE remains widely used in practical applications. This question represents an avenue for future research, which we leave it as future work.

\section{Conclusion}
We explored the impact of preconditioners on the stochastic training of neural fields, developing a theoretical framework based on Hessian-vector products. Our analysis demonstrates that curvature-aware preconditioners can significantly accelerate training for neural fields with non-traditional activations, such as sine, Gaussian, and wavelet. Experimental validation across various tasks—including image reconstruction, shape modeling, and neural radiance fields—using ESGD and AdaHessian confirmed these benefits. In contrast, for neural fields with ReLU or ReLU-PE activations, our theory predicted no significant advantage from curvature-aware preconditioning, highlighting the activation-dependent nature of these methods. For practitioners employing activations like sine, Gaussian, or wavelet, robust curvature-aware algorithms provide an efficient approach for stochastic training across diverse applications.

\clearpage
{
    \small
    \bibliographystyle{ieeenat_fullname}
    \bibliography{main}
}


\end{document}